\documentclass[letterpaper, 10 pt, conference]{ieeeconf}  

\IEEEoverridecommandlockouts                              

\usepackage{cite}
\usepackage{amsmath,amssymb,amsfonts}
\usepackage{algorithmic}
\usepackage{graphicx}
\usepackage{textcomp}
\usepackage{xcolor}
\usepackage{enumerate}
\usepackage[linesnumbered,ruled,lined]{algorithm2e}
\usepackage{booktabs}
\usepackage{threeparttable}
\usepackage{multirow}
\usepackage{stfloats}
\usepackage{subfigure}
\usepackage{graphicx}
\usepackage{rotating}
\usepackage{hyperref}

\title{\LARGE \bf
Multi-level Map Construction for Dynamic Scenes
}

\author{Xinggang Hu$^{1,2}$
 	\thanks{$^{1}$School of Electronic Information and Electrical Engineering, Dalian University of Technology, Dalian, China.}%
  	\thanks{$^{2}$College of Information Science and Engineering, Northeastern University, Shenyang, China.}%
}

\begin{document}

\maketitle
\thispagestyle{empty}
\pagestyle{empty}

\begin{abstract}
In dynamic scenes, both localization and mapping in visual SLAM face significant challenges. In recent years, numerous outstanding research works have proposed effective solutions for the localization problem. However, there has been a scarcity of excellent works focusing on constructing long-term consistent maps in dynamic scenes, which severely hampers map applications. To address this issue, we have designed a multi-level map construction system tailored for dynamic scenes. In this system, we employ multi-object tracking algorithms, DBSCAN clustering algorithm, and depth information to rectify the results of object detection, accurately extract static point clouds, and construct dense point cloud maps and octree maps. We propose a plane map construction algorithm specialized for dynamic scenes, involving the extraction, filtering, data association, and fusion optimization of planes in dynamic environments, thus creating a plane map. Additionally, we introduce an object map construction algorithm targeted at dynamic scenes, which includes object parameterization, data association, and update optimization. Extensive experiments on public datasets and real-world scenarios validate the accuracy of the multi-level maps constructed in this study and the robustness of the proposed algorithms. Furthermore, we demonstrate the practical application prospects of our algorithms by utilizing the constructed object maps for dynamic object tracking.
\end{abstract}

\section{INTRODUCTION}
The various geometric constraints that SLAM systems rely on are based on the assumption of a static environment. The introduction of dynamic factors makes these constraints difficult to hold. Regarding the visual SLAM problem in dynamic scenes, the current research efforts have mainly focused on localization, leading to the emergence of numerous excellent visual localization solutions\cite{bescos2018dynaslam,xiao2019dynamic,ji2021towards}. 

\begin{figure}[]
\centering
 	\includegraphics[scale=0.50]{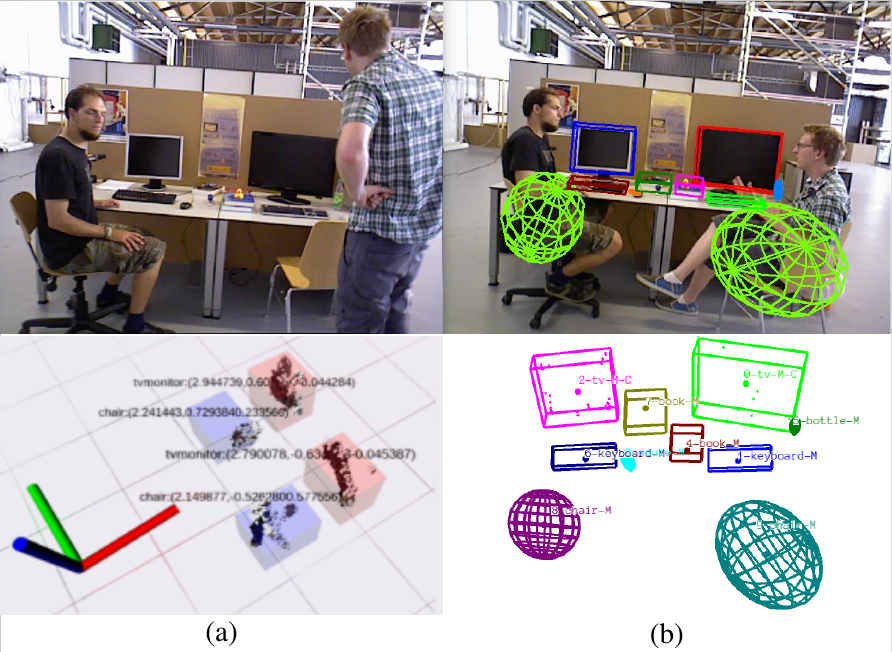}
 	\setlength{\abovecaptionskip}{-0.15cm} 
 	\caption{Object map construction results for dynamic scenes. (a) SG-SLAM; (b) Our algorithm.}
 	\label{Compared_with_SGSLAM}
 	\vspace{-7mm} 
 \end{figure}

However, apart from dealing with the impact of dynamic factors on camera localization, another challenge in visual SLAM for dynamic scenes lies in reconstructing the structure of the static environment. In contrast to static scenes, building maps for dynamic scenes faces significant challenges, primarily manifested in the following four aspects: 1) Dynamic factors can impact the accuracy of camera localization, consequently affecting mapping precision.
2) Establishing a long-term consistent map of the environment is more meaningful, but the introduction of dynamic factors reduces the reusability of the map. 3) Dynamic objects may irregularly occlude the static background, affecting map updates. 4) In dynamic scenes, there is a lack of sufficient viewpoint observations for static objects. Research on map construction for dynamic scenes is relatively scarce. Current research efforts mainly focus on building dense point cloud maps, dense maps, and octree maps after removing dynamic factors. Some works also incorporate semantic information into the constructed maps to enable higher-level map applications. For example, the latest research achievement addressing relevant issues, SG-SLAM\cite{cheng2022sg}, generates object maps in dynamic scenes as depicted in Fig.\ref{Compared_with_SGSLAM}(a). However, some issues remain: 1) Object modeling is rough and inaccurate, lacking complete and accurate parameterization of objects. 2) The semantic maps only contain few specific objects. These issues greatly limit the applications of the maps.

\begin{figure*}[]
\centering
 	\includegraphics[scale=0.305]{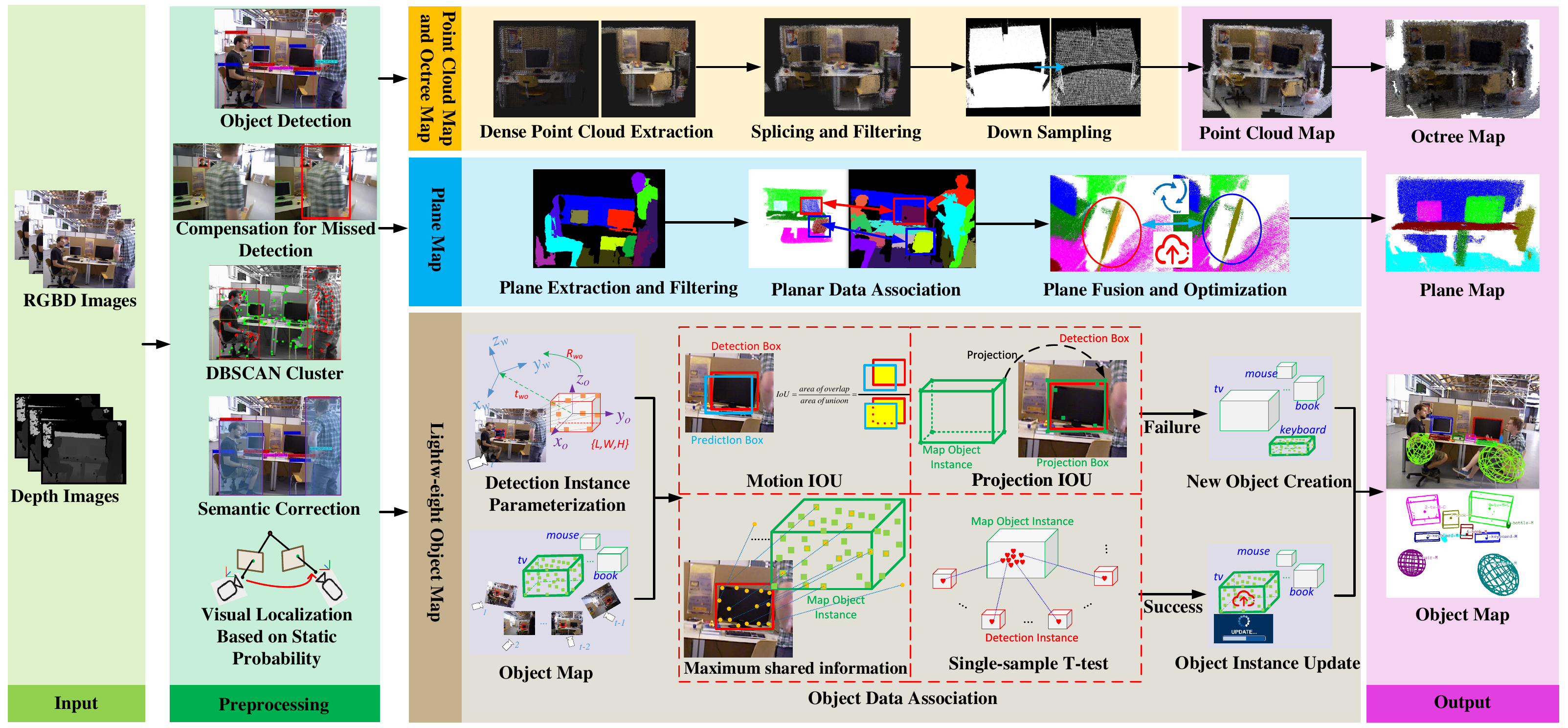}
 	\setlength{\abovecaptionskip}{-0.18cm} 
 	\caption{The System framework of multi-level map construction algorithm for dynamic scene. The light green part is the input module, which inputs RGB images and depth images. The dark green part is the preprocessing module, mainly responsible for obtaining and preprocessing semantic information. The yellow, blue and brown modules are mapping modules, which respectively represent the general process of building dense point cloud maps and Octree maps, plane maps, object maps. The purple red color is the output module, which outputs the multi-level map constructed by the mapping module.}
 	\label{Overview}
 	\vspace{-6mm} 
 \end{figure*}

In response to the above issues, this paper proposes a multi-level map construction algorithm for dynamic scenes, as shown in the system framework in Fig.\ref{Overview}. Firstly, YOLOX\cite{ge2021yolox} is utilized to obtain semantic information of the scene, and multi target tracking algorithm is employed for missed detection compensation, and the detected bounding boxes of potential moving objects are further refined using the DBSCAN density clustering algorithm and depth information. Subsequently, we extract point clouds and planes and utilize Principal Component Analysis (PCA) and minimum bounding rectangles to parameterize the objects. Additionally, we conduct filtering on the point clouds, planes, and objects. Next, based on the camera poses provided by our previous research\cite{hu2022cfp}, we perform point cloud stitching and fusion, and conduct data association and update optimization for planes and objects. Subsequently, the dense point cloud map is transformed into an octree map. Ultimately, this results in the construction of a multi-level map that includes dense point cloud maps, octree maps, plane maps, and object maps, thereby enriching the map's application scenarios. The effectiveness of our algorithms is fully validated through experiments conducted on publicly available datasets and real-world scenarios. As shown in Fig.\ref{Compared_with_SGSLAM}, compared to other state-of-the-art algorithms, the map constructed in this paper demonstrates a clear advantage in accuracy.

\textbf{The contributions of this paper are summarized as follows:}
\begin{itemize}
	\item We filter the point cloud based on the corrected object detection results to construct clean point cloud maps and octree maps that contain only static elements.
	\item We propose a method for building plane maps in dynamic scenes, enabling structural awareness of the environment.
	\item We propose a method for constructing object maps in dynamic scenes, allowing SLAM to serve higher-level requirements such as robot environment understanding, object manipulation, and semantic augmented reality.
    \item As far as we know, in dynamic scenes, this paper is the first work to construct plane maps, and also the first to accurately parameterize objects and build accurate and complete lightweight object maps. Additional experiments and videos are available on our project page \href{https://github.com/Hbelief1998/DyMLM-SLAM}{https://github.com/Hbelief1998/DyMLM-SLAM}.
\end{itemize} 

\section{Related work}

For dynamic scene mapping, many studies focus on constructing dense point cloud maps and octree maps. These studies primarily address two critical issues: first, tackling dynamic noise blocks arising from the absence of semantic information\cite{du2020accurate,xie2020moving,fan2022blitz,he2023ovd,ai2020ddl}; second, enhancing mapping efficiency\cite{yang2020sgc,ji2022robust,yang2019dre}. Additionally, other research\cite{scona2018staticfusion,palazzolo2019refusion,hasturk2021dudmap,zhu2022robust} has explored dense reconstruction based on voxels or TSDF. Dense mapping can cater to needs like augmented reality, but it is time-consuming. In contrast, plane-based mapping achieves similar functionality with lower computational requirements, but there is a lack of relevant research in this area.

Dense point cloud map, Octree map and dense map can clearly express the original appearance of static scene after removing dynamic objects, but can only meet the basic functions of mobile robot positioning and navigation, which limits the application scenarios of SLAM. If robots can perceive the surrounding environment at the conceptual level of humans, semantic information needs to be added to construct semantic maps, which has also been a research hotspot in recent years. DS-SLAM\cite{yu2018ds} used log-odds scoring to filter out unstable voxels and update the semantics of these voxels to construct a dense semantic 3D octree. Detect-SLAM\cite{zhong2018detect} propagates real-time movement probability for each key point and constructs an instance-level semantic graph of the environment. Han et al.\cite{han2020dynamic} used optical flow and PSPNet to detect and eliminate dynamic points, and simultaneously created semantic point cloud map and semantic Octree map. Wen et al.\cite{wen2021semantic} used depth error, photometric error and re-projection error to assign robust weights to static points, and deleted dynamic objects to build a Octree semantic map. Cheng et al.\cite{cheng2020improving} fused geometric information, semantic information, and human activity in an accurate and reliable 3D dense map. RS-SLAM\cite{ran2021rs} adoptd the maximum confidence fusion method for semantic updates, constructing a static background semantic OctoMap with semantic labels. SG-SLAM\cite{cheng2022sg} combined geometric and semantic information to rapidly remove dynamic features, then used the ROS interface to construct a semantic object map and a global octo-map. Yang et al.\cite{yang2022visual} combined the improved optical flow based mobile consistency detection method with the semantic segmentation results of depth information adjustment to build a semantic Octree map of static objects. However, the current semantic mapping scheme for dynamic scenes is rather rough in modeling static objects, and lacks the measurement of object pose, size and other information, and the general lack of object data association and the map of the object update, even if part of the work exists this link, the method is relatively simple, only suitable for simple scenes, lack of strong experimental support to prove the effectiveness of the method. In addition, the current research work mainly focuses on the chair and monitor two objects modeling, ignoring the other objects in the scene.


\section{Construction of geometric map} 

\subsection{Construction of PointCloud Map and Octree Map}\label{AA}

In the presence of semantic apriori, the point cloud in the target detection box or in the semantic mask can be deleted according to the semantic category to construct a dense point cloud map containing only static factors. However, relying solely on the original semantic results, the ``missed detection" and ``under-segmentation" issues of semantic information can result in incomplete removal of dynamic objects. In this paper, YOLOX\cite{ge2021yolox} is used for semantic information acquisition. To address the ``missed detection" problem, the paper utilizes multi target tracking algorithm\cite{bewley2016simple} for missed detection compensation. To address the ``under-segmentation" issue, we initially employ the DBSCAN clustering algorithm to extract foreground points within the bounding box of potential moving objects. Subsequently, we expand the detection box appropriately based on depth information from neighboring pixels along the detection box boundaries and the depth information of foreground points. To prevent errors caused by DBSCAN clustering, we set the expansion limit in all four directions of the detection box to 50 pixels.

In keyframes, pixels outside the corrected bounding boxes of potential moving objects are extracted and mapped to the 3D world coordinate system. Then, based on the camera poses provided by our previous research\cite{hu2022cfp}, the extracted point clouds from different keyframes are stitched and fused, followed by downsampling using voxel grid filtering. To enhance storage efficiency and support tasks like navigation and obstacle avoidance, the point cloud map is converted into an Octree map.

\subsection{Construction of Plane Map}\label{AA}
The PEAC algorithm\cite{feng2014fast} is used for plane extraction, obtaining the parameters and point cloud of the planes in the current camera coordinate system. The edge points of the planes are then extracted. Subsequently, PCL point cloud library is utilized for secondary fitting of the planes to obtain refined parameters and inliers, followed by outlier removal for the edge points of the planes. During this process, planes are filtered based on various factors such as depth info, inlier ratio and the positional relation with object detection boxes.

Once the plane map initialization is completed, the detected planes in the current frame and the existing planes in the map are traversed to establish data associations. Expressing the plane parameterization in the Hessian form $\pi = {({n^T},d)^T}$, where $n={{({{n}_{x}},{{n}_{y}},{{n}_{z}})}^{T}}$ represents the unit normal vector of the plane, and $d$ represents the distance from the plane to the origin. Let ${{P}_{c}}$ denote a plane instance detected in the current frame, with parameters ${{\pi }_{c}}={{({n}_{c}^{T},{{d}_{c}})}^{T}}$. Similarly, let ${{P}_{w}}$ represent a plane instance in the map, with parameters ${{\pi }_{w}}={{({n}_{w}^{T},{{d}_{w}})}^{T}}$.When $\text{Eq.(\ref{eq1})} \land (\text{Eq.(\ref{eq2})} \lor \text{Eq.(\ref{eq3})})$, the association is determined to be successful:

\vspace{-2mm}

\begin{equation}
\beta =\arccos (\frac{\left| {{({R}_{cw}^{-1}{{n}_{c}})}^{T}}\cdot {{n}_{w}} \right|}{\left| {R}_{cw}^{-1}{{n}_{c}} \right|\left| {{n}_{w}} \right|})<{{\beta }_{Th}}
    \label{eq1}
\end{equation}

\vspace{-2mm}

\begin{equation}
d=\left| t_{cw}^{T}{{n}_{c}}+{{d}_{c}}-{{d}_{w}} \right|<{{d}_{Th}}
    \label{eq2}
\end{equation}

\vspace{-2mm}

\begin{equation}
R=\frac{\left\| (n_{w}^{T}(R_{cw}^{-1}B_{c}^{i}-R_{cw}^{-1}{{t}_{cw}})+{{d}_{w}}<{{d}_{Th}}^{\prime })_{i=1}^{M} \right\|}{M}>{{R}_{Th}}
    \label{eq3}
\end{equation}

Where ${T_{cw}} \in {R^{4\times 4}}$ represents the transformation matrix from the world coordinate system to the current frame camera coordinate system. It includes the rotation matrix ${R_{cw}} \in {R^{3\times 3}}$ and the translation vector ${t_{cw}} \in {R^{3\times 1}}$. ${\beta_{Th}}$denotes the preset angle threshold. ${{\beta}_{Th}}$, ${{R}_{Th}}$, ${{d}_{Th}}^{\prime}$, and ${{d}_{Th}}$ respectively represent the angle threshold for normal vectors, the distance threshold between planes, the distance threshold from points to planes, and the threshold for the proportion of edge points satisfying the conditions. $M$ represents the number of edge points ${{B}_{c}}$ of plane ${{P}_{c}}$, $\left\| \centerdot  \right\|$ denotes the number of points satisfying the conditions, and $R$ represents the proportion of edge points satisfying the conditions.

After successful data association, update the point cloud, parameters, edge point cloud, and other relevant information of the map plane ${{P}_{w}}$ using the detected plane ${{P}_{c}}$.
The above strategy can generally achieve robust and accurate plane data association. However, in complex dynamic scenes, the detected planes often exhibit significant errors and randomness, leading to failures in plane data association. With more observations, the two planes with missed associations will optimize towards the right direction, making it easier to associate them later. Therefore, in the local mapping thread, a pairwise comparison is performed on the planes in the map. If two planes satisfy the association criteria mentioned above, they are considered as potentially missed associations. The plane with fewer observations is then merged into the one with more observations and optimized. Subsequently, the plane with fewer observations is removed from the map.


\section{Construction of Object Map} 
\subsection{Object Parameterization}\label{AA}
Since the objects to be modeled generally belong to the background and are located far from the camera, the extracted map points are often sparse in quantity and of poor quality, making it infeasible to perform outlier removal using clustering algorithms. Therefore, in this study, dense point clouds are employed for object modeling in each frame, and the DBSCAN density clustering algorithm is utilized to process the point clouds.

The object is parameterized as $O={R,t,s}$, which includes the rotation matrix ${{R}_{wo}}\in {{R}^{3\times 3}}$, the translation vector ${{t}_{wo}}\in {{R}^{3\times 1}}$, and the object dimensions $s=\{l,w,h\} \in {{R}^{3\times 1}}$. Let $P$ be the point cloud set of a certain object after being processed by the DBSCAN clustering algorithm, containing $m$ sample points. The coordinates of a sample point in the world coordinate system are denoted as $P_{i}^{w}={{(x_{i}^{w},y_{i}^{w},z_{i}^{w})}^{T}}$. Firstly, decentralize the sample. Let the center of the object point cloud set $P$ be denoted as $\bar{P}={{(\bar{x},\bar{y},\bar{z})}^{T}}$, and the decentralized point cloud set is represented as ${P}^{d}$. Get the three eigenvalues of the Covariance matrix of ${{P} ^ {d}}$, and the eigenvectors corresponding to the eigenvalues in descending order are the three principal axis directions of the object. The normalized eigenvectors ${{V}_{1}}$, ${{V}_{2}}$ and ${{V}_{3}}$ form a new coordinate axis ${R}_{wo}=({{V}_{1}}\text{ }{{V}_{2}}\ {{V}_{3}})$. Transforming ${P}^{d}$ to ${R}_{wo}$ yields $P_{o}^{d}$. By determining the extrema of $P_{o}^{d}$ along the three axes, the object's dimensions $s={l, w, h}$ and the geometric center of the 3D bounding box ${C}^{o}$ can be obtained. This enables the calculation of ${t}_{wo}$:

\vspace{-2mm}

\begin{equation}
{{t}_{wo}}={{C}^{w}}={{R}_{wo}}{{C}^{o}}+\bar{P}
    \label{eq6}
\end{equation}

\vspace{-2mm}

Thus, the object has been fully parameterized based on the PCA algorithm. When a certain plane's shape is close to a square, there are significant local clusters, or there are many outliers, it may not be suitable to use PCA alone to accurately estimate the three-dimensional point cloud's minimum bounding box. In such cases, the parameters obtained from the PCA algorithm can be corrected by using the minimum bounding rectangle.

Transforming the point cloud of the object in the world coordinate system to the object coordinate system:

\vspace{-2mm}

\begin{equation}
P_{i}^{o}={R}_{wo}^{T}P_{i}^{w}-{R}_{wo}^{T}{{t}_{wo}}
    \label{eq7}
\end{equation}

\vspace{-2mm}

Next, the point cloud data is projected onto the axes representing the length and width of the object, denoted as ${{(P_{i}^{o})}^{xy}}$. Then, compute the minimum bounding rectangle of the 2D point set ${{({{P}^{o}})}^{xy}}$ using the Graham scan algorithm and the rotating calipers algorithm. The two axis vectors of the minimum enclosing rectangle are denoted as ${{V}_{1}}^{\prime }=(V_{x1}^{\prime },V_{y1}^{\prime },0)^{\text{T}}$ and ${{V}_{2}}^{\prime }=(V_{x2}^{\prime },V_{y2}^{\prime },0)^{\text{T}}$. Then, these two axis vectors are transformed into the world coordinate system:

\vspace{-2mm}

\begin{equation}
{{V}_{i}}^{\prime \prime }={{R}_{wo}}{{V}_{i}}^{\prime }+{{t}_{wo}}, \quad i=1,2
    \label{eq8}
\end{equation}

\vspace{-2mm}

Normalize the matrix $({{V}_{1}}^{\prime \prime },{{V}_{2}}^{\prime \prime },{{V}_{3}})$ to obtain the rotation matrix ${{R}_{wo}}^{\prime }$ of the object coordinate system relative to the world coordinate system after the correction using the minimum enclosing rectangle.

\subsection{Object Data Association}

In the current frame $k$, for each detected instance $D_i^k$, we perform association judgment with each object instance $O_j^k$ in the map. Motion IoU, projection IoU, Maximum shared information, and single-sample t-test are common strategies for object data correlation. 
However, each of these strategies has limitations: Motion IoU relies on historical information to predict current frame detection boxes, which can be affected by missed and false detections; Projection IoU is susceptible to the phenomenon of object overlap in certain camera viewpoints caused by camera motion; Maximum shared information is prone to association failure when map points have poor quality in complex dynamic scenarios; The single-sample t-test has poor reliability when the number of observations of objects is small. These strategies can complement each other when integrated, leading to a more robust, accurate, and versatile object data correlation algorithm. 

The process of object data association based on single-sample t-test is as follows. The centroid cluster $C$ based on multi-frame observations of $O_{j}^{k}$ roughly follows a Gaussian distribution\cite{wu2020eao} $C \sim \mathcal{N}(\mu, \sigma^2)$, and we standardize it as follows:

\vspace{-2mm}

\begin{equation}
\frac{\mu (C)-{{c}_{d}}}{\sigma /\sqrt{m}}=\frac{\sqrt{m}(\mu (C)-{{c}_{d}})}{\sigma }\sim\mathcal{N}(0,1)
    \label{eq9}
\end{equation}

\vspace{-2mm}

Where $m$ is the number of samples in $C$, ${{c}_{d}}$ represents the centroid of $D_{i}^{k}$, and $\sigma$ is the population variance. The t-distribution is composed of a standard normal distribution and a $\chi^{2}$ distribution, where the degrees of freedom depend on the degrees of freedom of $\chi^{2}$ distribution:

\vspace{-2mm}

\begin{equation}
\frac{\mathcal{N}(0,1)}{\sqrt{{{\chi }^{2}}(m-1)/(m-1)}}\text{=}\frac{\sqrt{m}(\mu (C)-{{c}_{d}})}{\sigma (C)}\sim t(m-1)
    \label{eq10}
\end{equation}

\vspace{-1mm}

Hence, define t-statistic with $m-1$ degrees of freedom:

\begin{equation}
t = \frac{\mu(C) - {c}_{d}}{\sigma(C) / \sqrt{m}} \sim {t}(m-1)  
    \label{eq11}
\end{equation}

In order for the null hypothesis that $C$ and ${{c}_{d}}$ come from the same population to hold, the t-statistic should satisfy:

\vspace{-2mm}

\begin{equation}
f(t) \geq f\left(t_{\alpha / 2, v}\right)=\alpha / 2
    \label{eq12}
\end{equation}

\vspace{-2mm}

Where ${t}_{\alpha /2,v}$ is the upper $\alpha / 2$ quantile value of the t-distribution with $v\text{=}m-1$ degrees of freedom, if the t-statistic satisfies the Eq.\ref{eq12}, the two objects can be associated.

When $D_{i}^{k}$ and $O_{j}^{k}$ are determined to be associated through projection IoU, along with the use of either motion IoU, maximum shared information, or one-sample t-test, the final determination of successful association is made.

 \begin{figure}[h]
\vspace{-2mm}
\centering
 	\includegraphics[scale=0.60]{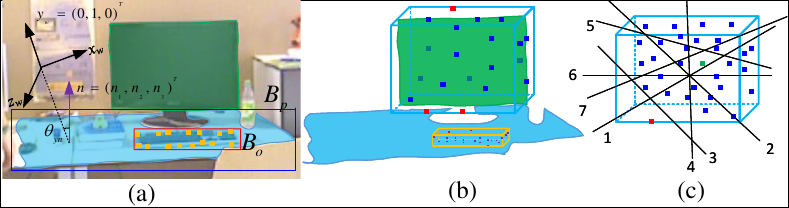}
 	\setlength{\abovecaptionskip}{-0.18cm} 
 	\caption{Mappoints outlier elimination. (a) Determine the desktop plane. (b) Remove outliers based on the distance from points to the plane. (c) Remove outliers using the Isolation Forest algorithm.}
 	\label{outlier elimination}
 	\vspace{-5mm}
 \end{figure}
 
\subsection{Object Update and Optimization}
We parametrize the detection instance and object instance using dense point clouds and sparse map points, respectively. This approach compensates for the deficiencies of insufficient map points in a single frame and the significant time consumption of dense point clouds in multiple frames. After successful data association, $O_{j}^{k}$'s map points and parameters are updated. Specifically, new map points are added to $O_{j}^{k}$. Afterwards, both the distance of the object's map points to the table plane or the plane associated with the object, and the Isolation Forest algorithm are utilized to remove outliers among these map points, as shown in Fig.\ref{outlier elimination}. After updating the map points, $O_{j}^{k}$ is re-parameterized. The decision to accept this update is based on two factors: the average IoU from projecting onto historical frames and the size prior.

 \begin{figure*}[b]
	\vspace{-5mm}
\centering
 	\includegraphics[scale=0.46]{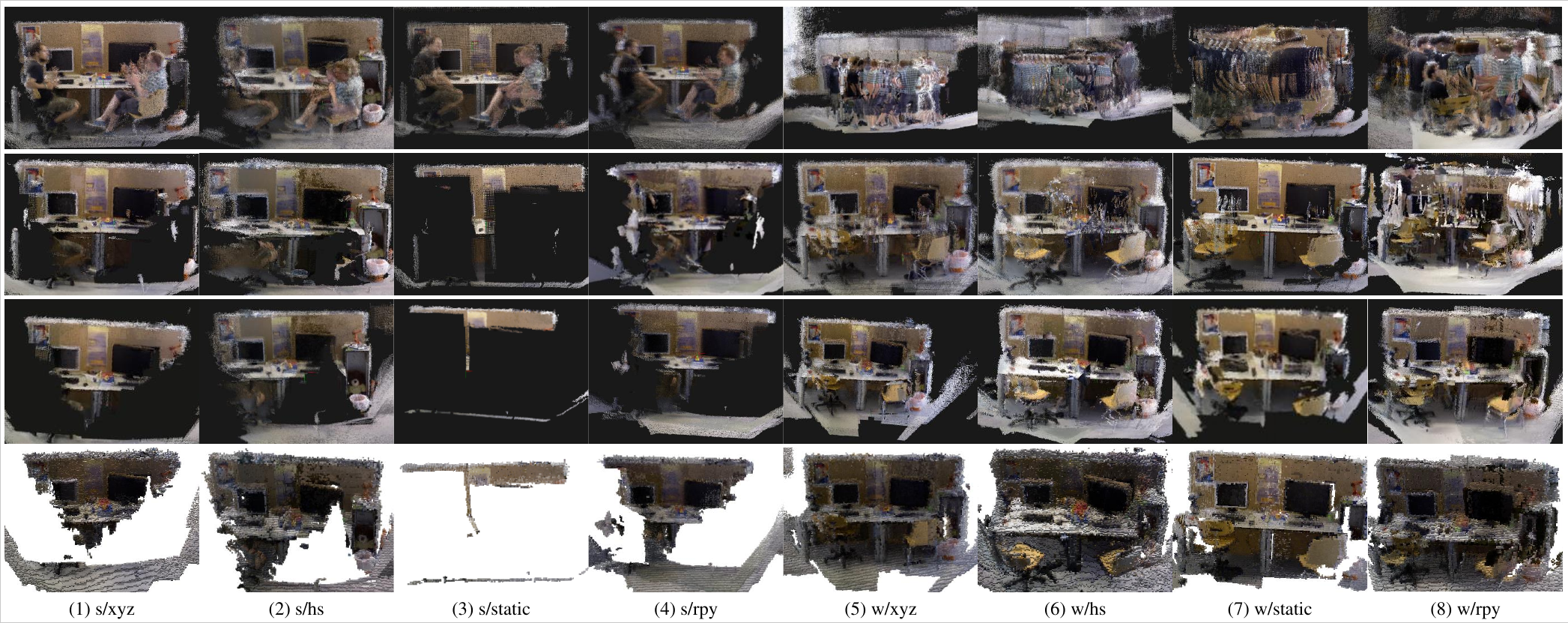}
 	\setlength{\abovecaptionskip}{-0.18cm} 
 	\caption{Point cloud map and Octree map. The top row displays the dense point cloud map built using the ORB-SLAM2 algorithm with the addition of the dense mapping module. The second row shows the dense point cloud map constructed using a previous study's method \cite{hu2022cfp} as the localization module, with the exclusion of point clouds within detected regions of potential moving objects. The third row exhibits the dense point cloud map constructed by our algorithm. The bottom row presents the octree map generated by our algorithm.}
 	\label{PCO-map}
 	\vspace{-3.5mm} 
 \end{figure*}

 \begin{figure*}[b]
\centering
 	\includegraphics[scale=0.46]{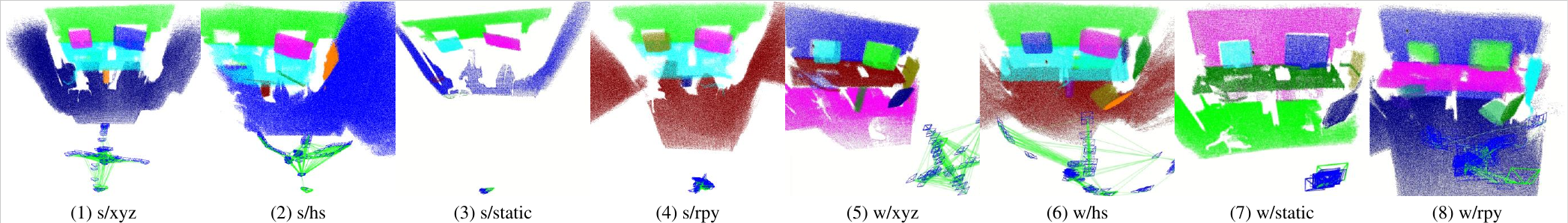}
 	\setlength{\abovecaptionskip}{-0.18cm} 
 	\caption{Plane map.}
 	\label{Plane-map}
 \end{figure*}

\section{Experiments and Evaluation}
We evaluated the performance of our algorithm on the TUM RGB-D dataset\cite{sturm2012benchmark} and in real-world scenarios, and applying the algorithm to dynamic object tracking. The TUM public dataset consists of four low-dynamic sequences (abbreviated as ``s/*") and four high-dynamic sequences (abbreviated as ``w/*"), where ``*" represents camera motions such as xyz, halfsphere, static, and rpy. In the low-dynamic sequences, two individuals are seated at a table and perform minor limb movements; in the high-dynamic sequences, two individuals move randomly within a room. The primary focus of this work is on mapping, and as ground truth maps are not available for the tested sequences, the experiments mainly aim to qualitatively demonstrate the mapping results. Our algorithm was executed on a laptop with an i9-12900H CPU, 3060 GPU, and 16GB of RAM.
\subsection{Construction of Geometric Map}\label{QSIM}
The construction results of dense point cloud map and octree map are shown in Fig.\ref{PCO-map}. It is observed that the ORB-SLAM2 algorithm fails to perform localization and mapping in high dynamic scenes due to the lack of a module to handle dynamic objects. In low dynamic scenes, the algorithm retains dynamic object point clouds. Due to missed detections in object detection and the challenge of fully covering potential moving objects with bounding boxes, the dense point cloud map constructed by removing point clouds within the original detection bounding boxes of potential moving objects contains numerous residual traces of these objects. 
In contrast, our algorithm benefits from compensating for missed detections and correcting the detection bounding boxes, resulting in a clear and clean dense point cloud map.
However, in the s/static sequence, the map construction is incomplete due to limited observation angles. Compared to the dense point cloud map, the octree map significantly reduces storage space and meets navigation and obstacle avoidance requirements. The constructed plane map in Fig.\ref{Plane-map} accurately perceives static background plane structures in dynamic scenes. This can be applied in advanced scenarios like augmented reality and serve as landmarks to enhance camera pose estimation accuracy.

\subsection{Construction of Object Map}
\begin{figure*}[t]
\centering
 	\includegraphics[scale=0.48]{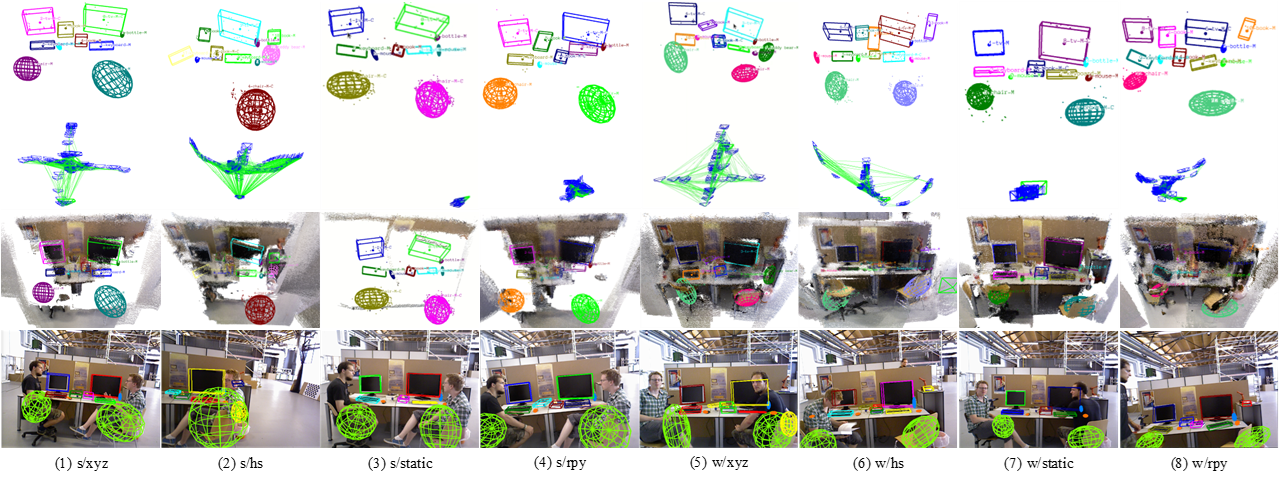}
 	\setlength{\abovecaptionskip}{-0.18cm} 
 	\caption{Object map. Regular-shaped objects, such as monitors, books, and keyboards, are represented using cubic boxes, while irregular-shaped objects, such as chairs, bottles, and teddy bears, are represented using quadric surfaces.}
 	\label{object-map}
 	\vspace{-3.5mm} 
 \end{figure*}

We evaluated the performance of object map construction on 8 dynamic sequences from the TUM dataset, as shown in Fig.\ref{object-map}. To verify the accuracy of object map construction, we overlay the constructed object models on the dense map and project them onto the image plane. In high-dynamic scenes, our algorithm accurately models almost all objects in the scene, unaffected by the camera's different motion patterns and the presence of dynamic objects in the environment. However, in low-dynamic scenes, where the two people are consistently sitting at the table, significant occlusion of static objects and the background occurs. As a result, our algorithm lacks sufficient observations for certain objects, leading to less accurate modeling for some objects, and this is unavoidable. The experimental results demonstrate the effectiveness of our algorithm in object parameterization, object data association, and object optimization strategies. By overcoming the influence of dynamic objects, the constructed object map provides strong support for subsequent applications such as semantic navigation, object grasping, and augmented reality.

 \begin{figure*}[t]
\centering
 	\includegraphics[scale=0.13]{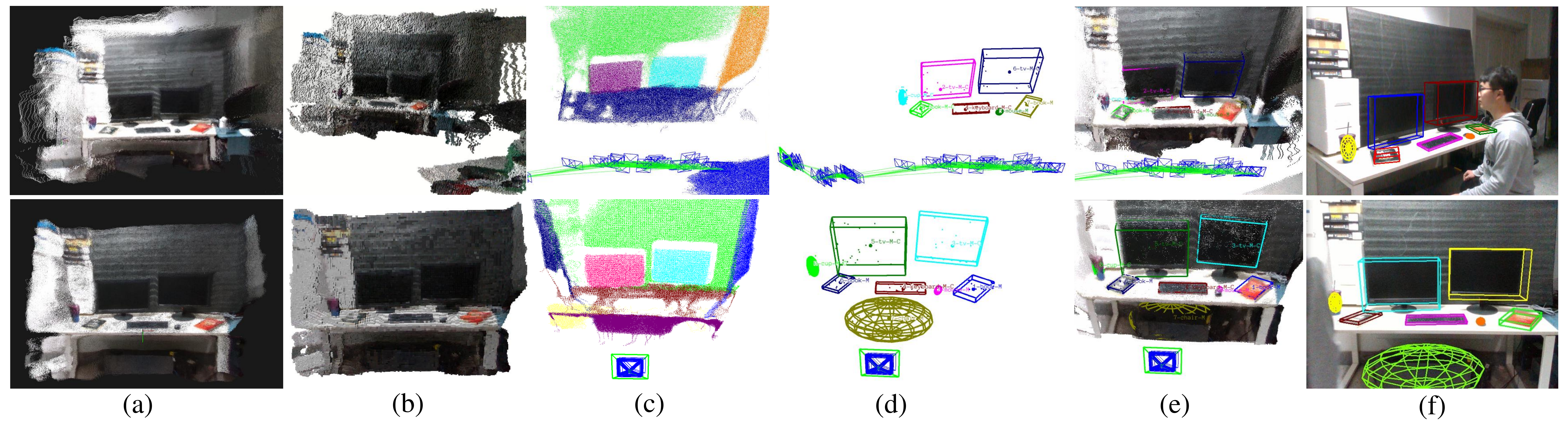}
 	\setlength{\abovecaptionskip}{-0.25cm} 
 	\caption{Multi-level map construction results in real-world scenarios. In the upper set of images, the camera moves from one end of the scene to the other; in the lower set of images, the camera remains almost stationary. (a), (b), and (c) represent dense point cloud map, octree map, and plane map, respectively. Image (d) shows the lightweight object map, where objects are overlaid on the dense point cloud map (Image (e)) and projected onto the image (Image (f)) to demonstrate the effectiveness of object map construction.}
 	\label{realsense-map}
 	\vspace{-7mm} 
 \end{figure*}
 
\subsection{Robustness Test in Real Environment}
We also tested our approach in real-world scenes using Realsense D435i camera to validate its effectiveness and robustness. In the experiments, a person performed irregular movements within the camera's field of view. To validate the algorithm's robustness, we evaluated two camera motions: 1) motion from one end of the scene to the other; 2) nearly still. The results of the multi-level map construction are shown in Fig.\ref{realsense-map}. The experimental results demonstrate that our algorithm can construct accurate dense point cloud maps, octree maps, plane maps, and lightweight object maps under different motion states of objects and the camera. 

\subsection{Dynamic Object Tracking Experiment}
\begin{figure}[h]
	\vspace{-3mm}
\centering
 	\includegraphics[scale=0.28]{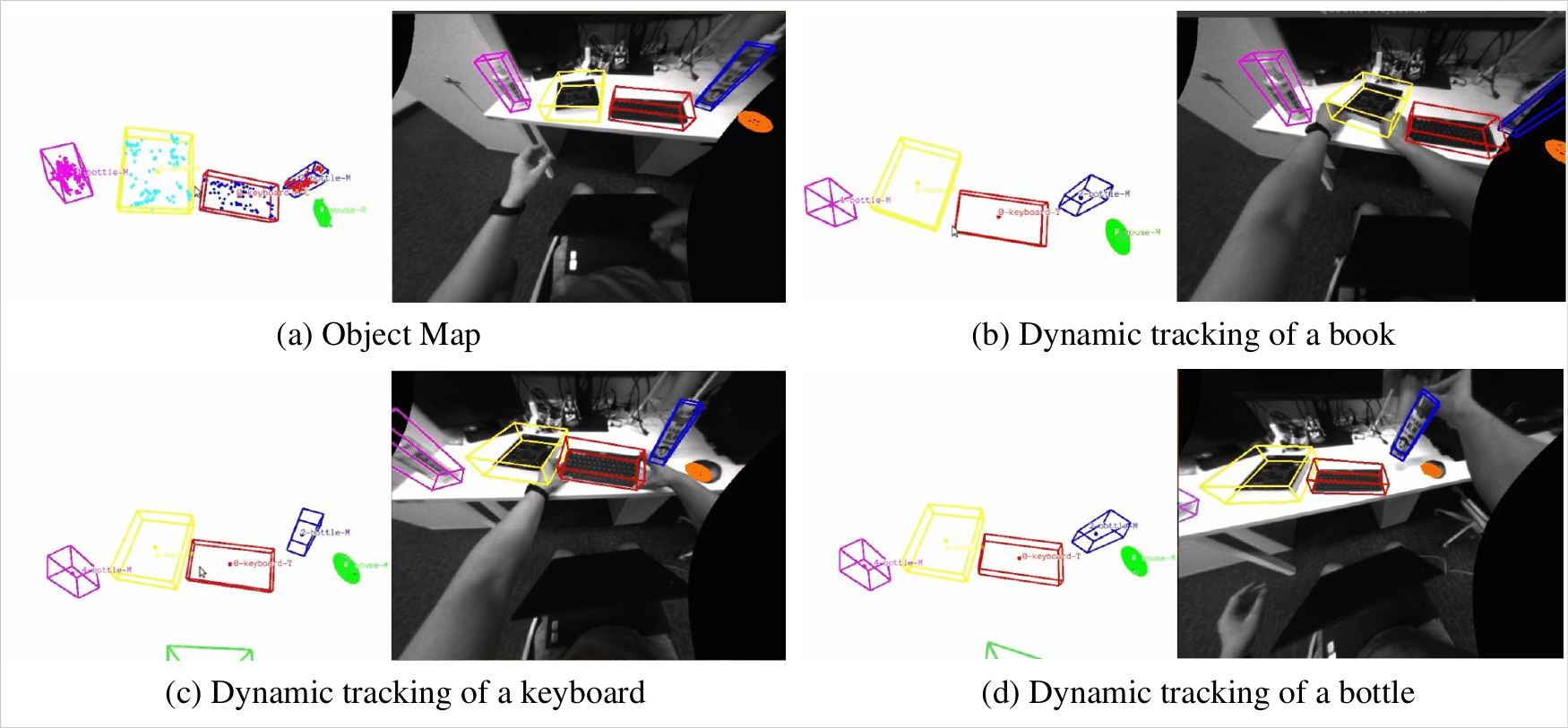}
 	\setlength{\abovecaptionskip}{-0.18cm} 
 	\caption{Object modeling and dynamic tracking in real-world scenarios.}
 	\label{DyObjTrack}
 	\vspace{-3.5mm} 
 \end{figure}

We further applied the constructed object maps to dynamic object tracking. We used the Pico Neo3 device to capture scene images and utilized our algorithm to build object maps. In this case, depth information of map points was obtained through stereo matching, with these computations performed only at keyframes to ensure real-time performance. The constructed object maps are shown in Fig.\ref{DyObjTrack}(a). Once the object map is built, users can select the target object for tracking. As the user moves the object, the system employs KCF single-object tracking and optical flow tracking algorithms to calculate the object's real-time pose. Fig.\ref{DyObjTrack}(b)-(d) demonstrate the dynamic tracking results for a book, keyboard, and bottle, respectively. The experimental results show that our algorithm can accurately model objects in dynamic environments, providing accurate object models and poses for object tracking, making it valuable for practical applications. Furthermore, this highlights that our algorithm is not device-specific, showcasing its robustness and versatility.

\section{Conclusion}

In this paper, we propose a multi-level map construction algorithm tailored to dynamic scenes. We successfully construct dense point cloud maps, octree maps, plane maps, and object maps containing only static backgrounds and objects in the presence of dynamic disturbances. This enriches the environmental perception capabilities of mobile robots and expands the application scenarios of maps built in dynamic environments. Extensive experiments demonstrate the accuracy and robustness of our algorithm, and the dynamic object tracking experiments further confirm its practicality. In the future, we plan to consider the real movements of other movable objects besides humans and optimize camera poses using planes and objects as landmarks to further enhance localization accuracy.


\bibliographystyle{IEEEtran}
\bibliography{hu2023}



\end{document}